\pdfoutput=1

\documentclass[11pt]{article}

\usepackage{acl}

\usepackage{times}
\usepackage{latexsym}

\usepackage[T1]{fontenc}

\usepackage[utf8]{inputenc}

\usepackage{microtype}

\usepackage{inconsolata}

\usepackage{subcaption}
\usepackage{graphicx}

\usepackage{booktabs}
\usepackage{multirow}

\usepackage{color}
\usepackage{xifthen}

\usepackage{enumerate}
\usepackage{enumitem}
\usepackage{tikz}
\usepackage{setspace}

\newcommand*\circled[1]{\tikz[baseline=(char.base)]{%
            \node[shape=circle,draw,inner sep=1pt] (char) {\bfseries\small#1};}}
\setlist[enumerate]{wide=0pt, leftmargin=0pt, listparindent=\parindent, itemsep=\parskip, parsep=0pt}

\newcommand{\hh}{\mymacro{\mathbf{h}}}
\newcommand{\bb}[1][]{\ifthenelse{\isempty{#1}}{\mymacro{\mathbf{b}}}{\mymacro{\mathbf{b}_{\text{#1}}}}}
\newcommand{\ff}[1][]{\ifthenelse{\isempty{#1}}{\mymacro{f}}{\mymacro{f_{\text{#1}}}}}
\newcommand{\qq}{\mymacro{\mathbf{q}}}
\newcommand{\kk}{\mymacro{\mathbf{k}}}
\newcommand{\vv}{\mymacro{\mathbf{v}}}
\newcommand{\HH}{\mymacro{\mathbf{H}}}
\newcommand{\FF}{\mymacro{\mathbf{F}}}
\newcommand{\GG}{\mymacro{\mathbf{G}}}
\newcommand{\xx}{\mymacro{\boldsymbol{x}}}

\newcommand{\yy}{\mymacro{\boldsymbol{y}}}

\newcommand{\W}[1][]{\ifthenelse{\isempty{#1}}{\mymacro{\mathbf{W}}}{\mymacro{\mathbf{W}_{\text{#1}}}}}

\newcommand{\calA}{\mymacro{\mathcal{A}}}
\newcommand{\aaa}{\mymacro{\mathbf{a}}}

\newcommand{\ww}{\mymacro{\boldsymbol{w}}}

\newcommand{\aalpha}{\mymacro{\boldsymbol{\alpha}}}
\newcommand{\e}{\mymacro{e}}

\newcommand{\stacktop}[1][]{\ifthenelse{\isempty{#1}}{\mymacro{\gamma^{\text{top}}}}{\mymacro{\gamma^{\text{top}}_{#1}}}}
\newcommand{\syma}{\mymacro{\texttt{a}}}
\newcommand{\symb}{\mymacro{\texttt{b}}}
\newcommand{\symc}{\mymacro{\texttt{c}}}
\newcommand{\symM}{\mymacro{\texttt{M}}}
\newcommand{\symP}{\mymacro{\texttt{P}}}

\newcommand{\bool}{\mymacro{\mathbb{B}}}

\newcommand{\alphabet}[1][]{\ifthenelse{\isempty{#1}}{\mymacro{\Sigma}}{\mymacro{\Sigma_{\text{#1}}}}}
\newcommand{\maskalphabet}{\mymacro{\widetilde{\alphabet}}}

\newcommand{\push}{\mymacro{\texttt{PUSH}}}
\newcommand{\pop}{\mymacro{\texttt{POP}}}
\newcommand{\noop}{\mymacro{\texttt{NO-OP}}}
\newcommand{\peek}{\mymacro{\texttt{PEEK}}}
\newcommand{\onehot}[1][]{\ifthenelse{\isempty{#1}}{\mymacro{ \llbracket \rrbracket}}{\mymacro{ \llbracket#1\rrbracket}}}

\newcommand{\tquery}{\mymacro{\text{(query)}}}
\newcommand{\tkey}{\mymacro{\text{(key)}}}
\newcommand{\tvalue}{\mymacro{\text{(value)}}}

\newcommand{\none}{\mymacro{\texttt{none}}}
\newcommand{\sincos}{\mymacro{\texttt{sincos}}}
\newcommand{\relative}{\mymacro{\texttt{relative}}}
\newcommand{\rotary}{\mymacro{\texttt{rotary}}}
\newcommand{\ALiBi}{\mymacro{\texttt{ALiBi}}}

\newcommand{\FFN}{\mymacro{\text{FFN}}}
\newcommand{\MHA}{\mymacro{\mathbf{M}}}
\newcommand{\LN}{\mymacro{\text{LN}}}
\newcommand{\Attention}{\mymacro{\mathbf{A}}}
\newcommand{\Stack}{\mymacro{\mathbf{S}}}
\newcommand{\Embedding}{\mymacro{\text{Embedding}}}
\newcommand{\PE}{\mymacro{\text{PE}}}
\newcommand{\mask}{\mymacro{\textsc{[mask]}}}
\newcommand{\eos}{{\mymacro{\textsc{[eos]}}}}
\newcommand{\bos}{{\mymacro{\textsc{[bos]}}}}
\newcommand{\pad}{\mymacro{\textsc{[pad]}}}
\newcommand{\tpusha}{\mymacro{[\textsc{push}\ \syma]}}
\newcommand{\tpushb}{\mymacro{[\textsc{push}\ \symb]}}
\newcommand{\tpop}{\mymacro{[\textsc{pop}]}}

\newcommand{\stackop}{\mymacro{\upsilon}}

\makeatletter
\newcommand*{\bigcdot}{}%
\DeclareRobustCommand*{\bigcdot}{%
  \mathbin{\mathpalette\bigcdot@{}}%
}
\newcommand*{\bigcdot@scalefactor}{.5}
\newcommand*{\bigcdot@widthfactor}{1.15}
\newcommand*{\bigcdot@}[2]{%
  \sbox0{$#1\vcenter{}$}%
  \sbox2{$#1\cdot\m@th$}%
  \hbox to \bigcdot@widthfactor\wd2{%
    \hfil
    \raise\ht0\hbox{%
      \scalebox{\bigcdot@scalefactor}{%
        \lower\ht0\hbox{$#1\bullet\m@th$}%
      }%
    }%
    \hfil
  }%
}
\makeatother

\usepackage{inconsolata}
\usepackage{multirow}

\usepackage{tikz}
\usetikzlibrary{shapes.multipart}
\usetikzlibrary{arrows}
\def\drawstack(#1, #2){
  \begin{tikzpicture}[baseline=0]
    \def\scale{0.6}
    \def\array{#1}
    \def\length{#2}
    \foreach \i/\j in \array {
      \node[draw, fill=lightgray!40, minimum size=\scale cm] at (\i * \scale, 0) {\j};
    }
    \fill [fill=lightgray] (\length * \scale + \scale * 0.5, 0 - \scale * 0.5) -- +(\scale, 0) -- +(\scale, \scale) -- +(0, \scale) -- cycle;
    \draw (\length * \scale + \scale * 0.5, 0 - \scale * 0.5) -- +(\scale, 0);
    \draw (\length * \scale + \scale * 0.5, 0 + \scale * 0.5) -- +(\scale, 0);
    \draw (\length * \scale + \scale * 0.5, 0 - \scale * 0.5) -- +(0, \scale);
  \end{tikzpicture}
}
\def\drawattention(#1, #2){
  \begin{tikzpicture}[baseline=-5]
    \def\scale{0.6}
    \def\array{#1}
    \def\top{#2}
    \foreach \i/\j in \array {
        \node[minimum size=\scale cm, align=center] at (\i * \scale, 0) (index\i) {\j};
    }
    \node at (\top* \scale, -\scale * 1.2) (pointer) {};
    \draw [-stealth] (pointer) -- (index\top);
  \end{tikzpicture}
}

\usepackage{listings}
\usepackage[most]{tcolorbox}

\usepackage{xparse}
\def\exampletext{Example} %

\NewDocumentEnvironment{testexample}{ O{} }
{
\colorlet{colexam}{red!55!black} %
\newtcolorbox[use counter=testexample]{testexamplebox}{%
    empty,%
    title={\exampletext: #1},%
    attach boxed title to top left,
       minipage boxed title,
    boxed title style={empty,size=minimal,toprule=0pt,top=4pt,left=3mm,overlay={}},
    coltitle=colexam,fonttitle=\bfseries,
    before=\par\medskip\noindent,parbox=false,boxsep=0pt,left=3mm,right=0mm,top=2pt,breakable,pad at break=0mm,
       before upper=\csname @totalleftmargin\endcsname0pt, %
    overlay unbroken={\draw[colexam,line width=.5pt] ([xshift=-0pt]title.north west) -- ([xshift=-0pt]frame.south west); },
    overlay first={\draw[colexam,line width=.5pt] ([xshift=-0pt]title.north west) -- ([xshift=-0pt]frame.south west); },
    overlay middle={\draw[colexam,line width=.5pt] ([xshift=-0pt]frame.north west) -- ([xshift=-0pt]frame.south west); },
    overlay last={\draw[colexam,line width=.5pt] ([xshift=-0pt]frame.north west) -- ([xshift=-0pt]frame.south west); },%
    }
\begin{testexamplebox}}
{\end{testexamplebox}\endlist}

\usepackage{custom}
\newcommand{\mymacro}[1]{{\color{black} #1}}

\newcommand{\defn}[1]{\textbf{#1}}

\newcommand{\paroutline}[3][false]{%
    \ifnum\pdfstrcmp{#1}{true}=0
        #3%
    \else
        [\textit{\textcolor{DiverseMagenta}{#2}}] \textcolor{AccentBlue}{#3}%
    \fi
}

\newcommand{\R}{{\mymacro{ \mathbb{R}}}}

\newcommand{\eosalphabet}{{\mymacro{ \overline{\alphabet}}}}

\newcommand{\stacksym}{{\mymacro{\stacksymbol{\gamma}}}}

\newcommand{\defeq}{\mathrel{\stackrel{\textnormal{\tiny def}}{=}}}

\newcommand{\stack}{{\mymacro{\textsf{stack}}}}

\newcommand{\negterm}[1]{{\mymacro{ {\raise.17ex\hbox{$\scriptstyle\sim$}} #1}}}

\newcommand{\stackseq}{{\mymacro{ {\boldsymbol{\gamma}}}}}

\newcommand{\stacksymbol}[1]{{\mymacro{ #1 }}}

\newcommand{\ignore}[1]{}
\newcommand{\expandLater}[1]{}

\def\1{\mathbf{1}}

\def\vv{{{\mymacro{ \mathbf{v}}}}}

\newcommand{\softmax}{{\mymacro{ \mathrm{softmax}}}}

\newcommand{\bigO}[1]{{\mymacro{ \mathcal{O}\left(#1\right)}}}

\newcommand{\ethz}{1}
\newcommand{\ucambridge}{2}

\title{A Transformer with Stack Attention}

\author{Jiaoda Li$^{\ethz}$ \quad 
Jennifer C. White$^{\ucambridge}$ \quad
Mrinmaya Sachan$^{\ethz}$ \quad 
Ryan Cotterell$^{\ethz}$ \quad 
\\
$^{\ethz}$ETH Zürich~\;~ $^{\ucambridge}$University of Cambridge \\
\{\texttt{\href{mailto:jiaoda.li@ai.ethz.ch}{jiaoda.li}},
\texttt{\href{mailto:mrinmaya.sachan@inf.ethz.ch}{mrinmaya.sachan}},
\texttt{\href{mailto:ryan.cotterell@inf.ethz.ch}{ryan.cotterell}}\}\texttt{@inf.ethz.ch} \quad \texttt{\href{mailto:jw2088@cam.ac.uk}{jw2088@cam.ac.uk}}
}

\begin{document}
\maketitle
\begin{abstract}
  Natural languages are believed to be (mildly) context-sensitive.
  Despite underpinning remarkably capable large language models, transformers are unable to model many context-free language tasks.
  In an attempt to address this limitation in the modeling power of transformer-based language models, we propose augmenting them with a differentiable, stack-based attention mechanism.
  Our stack-based attention mechanism can be incorporated into any transformer-based language model and adds a level of interpretability to the model.
  We show that the addition of our stack-based attention mechanism enables the transformer to model some, but not all, deterministic context-free languages.

\vspace{0.5em}
\hspace{.5em}\includegraphics[width=1.25em,height=1.15em]{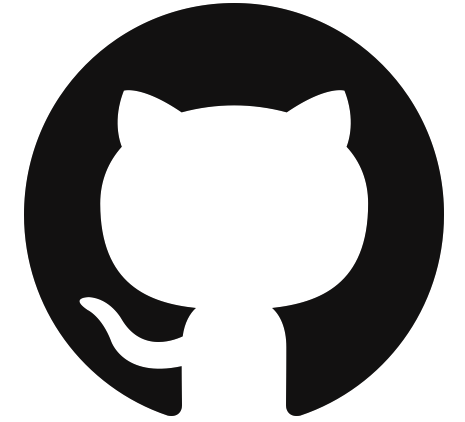}\hspace{.75em}
\parbox{\dimexpr\linewidth-7\fboxsep-7\fboxrule}{\url{https://github.com/rycolab/stack-transformer}}
\vspace{-.5em}
\end{abstract}

\section{Introduction}
Language models (LMs) based on the transformer architecture \citep{NIPS2017_3f5ee243} have shown great empirical success at a wide range of NLP tasks \cite{devlin-etal-2019-bert, radford2019language, liu2020roberta, NEURIPS2020_1457c0d6}.
However, recent theoretical \citep{hahn-2020-theoretical, angluin2023masked} and empirical \citep{ebrahimi-etal-2020-self, bhattamishra-etal-2020-ability, deletang2023neural} research suggests that language models based on transformers show difficulty in learning basic algorithmic patterns.
A prime example is the Dyck-$n$ language, i.e., the language of balanced parentheses of depth $\leq n$. 
When $n>1$, it has been argued that transformers are theoretically  \cite{hahn-2020-theoretical} and empirically \cite{ebrahimi-etal-2020-self} unable to learn a Dyck-$n$ language. 
Additionally, \citet{deletang2023neural} report that transformer-based LMs fail to learn four deterministic context-free (DCF) tasks.
The authors of this work contend that the resolution of this insufficiency is paramount if human-level language understanding is to be achieved by computers. 
Indeed, \citet{1056813} famously argues that human language has many context-free traits; see also \citet{CHOMSKY1963118}.
Moreover, \citet{Shieber1987} goes further and argues that snippets of Swiss German are even higher on the Chomksy hierarchy.\looseness=-1

The scientific question treated in this paper is whether there exists a minimal modification to the transformer architecture that \emph{does} allow it to learn a larger swathe of the formal languages most closely associated with human language.
Specifically, in this paper, we augment the transformer architecture with a novel stack attention mechanism that enables it to learn certain CF languages empirically.
Our stack attention mechanism simulates a stack by maintaining a probability distribution over which of the subsequently observed tokens is at the top element of the stack.
In turn, this probability distribution serves as an attention mechanism.
Compared to \citet{dusell2023stack}, which also applies stack augmentation to the transformer, our stack attention is more space efficient and allows for easier interpretation through visualizing the attention weights. 
We incorporate our innovation into the transformer by adding a stack attention sub-layer to each layer, rather than completely replacing the standard attention. 
Augmenting models in a modular way like this allows for direct integration with pre-trained transformer-based LMs.

We evaluate our stack-augmented transformer through comparison with a standard transformer on four DCF tasks taken from \citet{deletang2023neural}. 
We find that the stack-augmented transformer performs substantially better than the standard transformer on two of the four DCF tasks. 
Nevertheless, in contrast to \citet{dusell2023stack}, who claim their architecture can recognize the entire class of CF languages, we find transformers with our stack attention still struggle on two DCF tasks that involve modular arithmetic.

\section{Preliminaries}
In this section, we provide the necessary technical background for our exposition. 
We first review the self-attention mechanism and then introduce the transformer architecture.

\subsection{The Self-Attention Mechanism}
The attention mechanism \cite{DBLP:journals/corr/BahdanauCB14} is the fundamental building block of the transformer architecture \citep{NIPS2017_3f5ee243}, which we discuss in the next section.
One common form of attention is \defn{self-attention} \citep{cheng-etal-2016-long, parikh-etal-2016-decomposable}.
Our construction of a stack-augmented attention mechanism is a modification of this self-attention mechanism.

The premise of self-attention is as follows.
A sentence representation $\HH=[\hh_1; \ldots; \hh_N] \in \R^{D\times N}$ is a horizontal concatenation of column vectors $\hh_n$ in $\R^D$, where each column is a representation of the $n^{\text{th}}$ word.
Our goal is to construct a distribution over the index set $\{1, \ldots, N\}$, denoted as $[N]$.
We do so in three steps, described below.
\begin{enumerate}[label=\protect\circled{\arabic*}, itemsep=0.3cm]
\item The first step is to construct a real-valued, pair-wise compatibility score.
The most common way to do this is through a (scaled) dot-product, i.e., 
\begin{equation}
\label{eq:align}
    \e_{ij} \defeq \frac{\hh_i \bigcdot \hh_j}{\sqrt{D}}
\end{equation}
\item The second step is to take the pair-wise compatibility scores and project them onto the simplex $\Delta^{N-1}$ through the softmax.
This results in the following distribution
\begin{equation}
    \aalpha_i(j) \defeq \frac{\exp(\e_{ij})}{\sum_{n=1}^{N} \exp(\e_{in})}
\end{equation}
which is termed the \defn{self-attention distribution}.
Note there are $N$ self-attention distributions $\aalpha_i$.
\item The third, and final, step is to construct a weighted average of the representations $\HH=[\hh_1; \ldots; \hh_N] \in \R^{D\times N}$ using the self-attention distribution as follows\looseness=-1
\begin{equation}
\label{eq:attention}
    \Attention(\HH)_{:,i} \defeq \sum_{n=1}^{N} \aalpha_i(n)\,\hh_n
\end{equation}
where $\Attention(\HH)_{:,i}$ denotes the $i^{\text{th}}$ column of $\Attention(\HH)$.
The function $\Attention : \R^{D \times N} \rightarrow \R^{D \times N}$ (for any $N$), as defined above, is called an \defn{attention head}.
\end{enumerate}
Importantly, we see that $\Attention$ is a differentiable function.
Differentiability allows us to learn the parameters of an attention head with gradient-based methods. 
And, more importantly, it has a specific desirable property---namely, it is invariant with respect to permutations of the columns of $\HH$.
Computationally, this implies that $\Attention(\HH)_{:,i}$ and $\Attention(\HH)_{:,j}$ can be computed in parallel for $i \neq j$.
It is specifically this form of parallelism that grants the transformer architecture its ability to scale.
One drawback of the permutation invariance, however, is that $\Attention$ is not a linguistically plausible mechanism as human language is decidedly not permutation invariant.  
This problem is addressed through the incorporation of attention masks and positional encodings \citep[\S~3.5]{NIPS2017_3f5ee243} in the transformer architecture, as we discuss in \Cref{sec:transformer-architecture}.\looseness=-1

\paragraph{Masked Self-Attention.}
An attention mask $\GG\in\bool^{N\times N}$, where $\bool = \{0, 1\}$, can be applied to the self-attentions using the following generalization
\begin{equation}
    \aalpha_i(j) = \frac{\exp(\e_{ij})\GG_{i,j}}{\sum_{n=1}^{N} \exp(\e_{in}) \GG_{i,n}}
\end{equation}
Attention masks allow for hard constraints on which indices can be attended to by the attention head. 
A commonly used masking scheme is \defn{future masking} where each position is only allowed to attend to positions up to and including itself, i.e., we define the following mask
\begin{equation}
    \label{eq:mask}
    \GG_{i,n} = 
    \begin{cases}
        1, &  n <  i  \\
        0, &  n \geq i
    \end{cases}
\end{equation}
Future masking allows transformers to be used in autoregressive language models by preventing the model from peeking at words that have yet to be generated, which we detail in \cref{sec:probability-models}.

\paragraph{Queries, Keys, and Values.}
In the version of the attention mechanism introduced by \citet{NIPS2017_3f5ee243}, the attention mechanism is augmented with additional linear projections. 
Specifically, the vectors $\hh_n$ are linearly projected to construct three new vectors, defined below
\begin{subequations}
\begin{align}
\qq_n & \defeq \W[Q] \hh_n  \quad\quad \quad \quad \quad \tquery  \\
 \kk_n& \defeq \W[K] \hh_n  \quad \quad \quad   \quad  \quad \tkey \\
  \vv_n& \defeq \W[V] \hh_n  \quad\quad \quad    \quad \quad \tvalue
\end{align}
\end{subequations}
where $\W[V], \W[Q], \W[K]\in\mathbb{R}^{D' \times D}$
are parameter matrices. 
Compatibility scores are then computed between the corresponding query--key pair:
\begin{equation}
    \e_{ij} = \frac{\qq_i \bigcdot \kk_j}{\sqrt{D'}}
\end{equation}
Using those compatibility scores, a self-attention distribution is constructed using the softmax.
Finally, as before, a weighted sum of the values is computed using the self-attention distribution:
\begin{equation}
    \Attention(\HH)_{:,i} = \sum_{n=1}^{N} \aalpha_i(n)\, \vv_n
\end{equation}

\paragraph{Multi-head Self-Attention.}
We additionally define the multi-head self-attention mechanism.
In \defn{multi-head attention}, we combine $M$ attention heads $\Attention^{(1)}, \ldots, \Attention^{(M)}$ as follows
\begin{equation}
    \MHA(\HH)_{:,i} \defeq \sum_{m=1}^M \W[O]^{(m)} \Attention^{(m)}(\HH)_{:,i}
\end{equation}
where $\W[O]^{(m)}\in\R^{D\times D'}$
is the output projection matrix for head $\Attention^{(m)}$. 
Usually, we set $D'=D/M$.

\subsection{The Transformer 
Architecture}\label{sec:transformer-architecture}
We now describe the transformer architecture. 
A transformer over a vocabulary $\alphabet$ constitutes a function of type\footnote{Type-theoretically, $N$ is a parameter of the type.
Thus, our type signature is a dependent type \citep{hottbook}
} $\alphabet^N \rightarrow \R^{D \times N}$  
where a string $\ww=w_1 \cdots w_{N}\in\alphabet^N$ of length $N$ is encoded into a $\R^{D \times N}$ representation where $D$ is the model size.
The transformer is defined compositionally over a sequence of layers.
First, we define
\begin{equation}
\HH^{(0)} \defeq \Embedding + \PE
\end{equation}
where $\Embedding$ of type $\alphabet^N \rightarrow \R^{D\times N}$
is the embedding layer and $\PE$ of type $\alphabet^N \rightarrow \R^{D\times N}$
is the positional encoding that injects information about the relative or absolute position of tokens, which extinguishes the permutation invariance of the transformer. 
Each transformer layer consists of two sub-layers: a multi-head self-attention $\MHA$ of type $\R^{D\times N} \rightarrow \R^{D\times N}$
and a fully connected feed-forward network $\FFN$ of type $\R^{D\times N} \rightarrow \R^{D\times N}$. A residual connection is employed around each sub-layer, followed by a layer normalization \citep{ba2016layer} $\LN$ of type $\R^{D\times N} \rightarrow \R^{D\times N}$
: for $0 < \ell \leq L$, we have the following recursive definition
\begin{subequations}
\begin{align}
    \HH^{(\ell)}_{\MHA} &\defeq \LN\left(\MHA\left(\HH^{(\ell-1)}\right) + \HH^{(\ell-1)}\right) \label{eq:translayer}\\
       \HH^{(\ell)}_{\FFN} &\defeq \LN\left(\FFN\left(\HH^{(\ell)}_{\MHA}\right) + \HH^{(\ell)}_{\MHA}\right) \label{eq:translayerffn} \\
    \HH^{(\ell)} &\defeq \HH^{(\ell)}_{\FFN} 
\end{align}
\end{subequations}
where $\HH^{(\ell)}_{\MHA}$, $\HH^{(\ell)}_{\FFN}$ and $\HH^{(\ell)}$ are functions of type $\alphabet^N \rightarrow \mathbb{R}^{D \times N}$ for any $N$. They have $\ww$ as input, which we omit for brevity when the context is clear.

\paragraph{Future-masked Transformer.}
If the future mask in \cref{eq:mask} is used in every $\HH^{(\ell)}_{\MHA}$, we call such a transformer \defn{future-masked transformer}, denoted as $\FF^{(L)}$.
As we will see, future-masked transformers are necessary to construct autoregressive language models, which cannot peek at the future.\looseness=-1

\begin{figure*}
    \centering
    \begin{tabular}{llll}
    \toprule
    Action & Stack & Attention & Stack Attention $\aalpha$ \\
    \midrule
     &\drawstack({}, 0) &\drawattention({1/$\bos$, 2/$\syma$, 3/$\symb$, 4/$\symc$}, 1) & $\aalpha_0 = [1, 0, 0, 0]^\top$ \\
    \push\ \syma & 
    \drawstack({1/$\syma$}, 1) & 
    \drawattention({1/$\bos$, 2/$\syma$, 3/$\symb$, 4/$\symc$}, 2) &
    $\aalpha_1 = [0, 1, 0, 0]^\top$
    \\
    \push\ \symb &
    \drawstack({1/$\symb$, 2/$\syma$}, 2) &
    \drawattention({1/$\bos$, 2/$\syma$, 3/$\symb$, 4/$\symc$}, 3) &
    $\aalpha_2 = [0, 0, 1, 0]^\top$
    \\
    \push\ \symc  &
    \drawstack({1/$\symc$, 2/$\symb$, 3/$\syma$}, 3) &
    \drawattention({1/$\bos$, 2/$\syma$, 3/$\symb$, 4/$\symc$}, 4) &
    $\aalpha_3 = [0, 0, 0, 1]^\top$
    \\
    \pop  &
    \drawstack({1/$\symb$, 2/$\syma$}, 2) &
    \drawattention({1/$\bos$, 2/$\syma$, 3/$\symb$, 4/$\symc$}, 3) &
    $\aalpha_4 = \sum_{j=1}^3 \aalpha_3(j) \aalpha_{j-1} = \aalpha_2= [0, 0, 1, 0]^\top$
    \\
    \noop  &
    \drawstack({1/$\symb$, 2/$\syma$}, 2) &
    \drawattention({1/$\bos$, 2/$\syma$, 3/$\symb$, 4/$\symc$}, 3) &
    $\aalpha_5 = \aalpha_4 = [0, 0, 1, 0]^\top$
    \\
    \pop  &
    \drawstack({1/$\syma$}, 1) &
    \drawattention({1/$\bos$, 2/$\syma$, 3/$\symb$, 4/$\symc$}, 2) &
    $\aalpha_6 = \sum_{j=1}^5 \aalpha_5(j) \aalpha_{j-1} = \aalpha_1 = [0, 1, 0, 0]^\top$
    \\
    \bottomrule
    \end{tabular}
    \caption{An example illustrating how attentions can emulate stacks. The first column lists the operation performed at each timestep. The second column presents the stack contents after performing the operation. The third column shows a hard attention over the input tokens. The pointer of the attention indicates the current stack top. The last column is the proposed stack attention.}
    \label{fig:stack}
\end{figure*}
\subsection{Probability Models}
\label{sec:probability-models}
Next, we describe two natural ways of constructing a probability distribution from a transformer.
{\paragraph{Masked Language Modeling.}
First, we consider the case of masked language modeling (MLM).
Masked language models perform the cloze task, i.e., they fill in a missing word given a left and right context.
More formally, consider a string $\ww \in \alphabet^*$ of length $T$.
Let $w_t$ denote the $t^{\text{th}}$ symbol in $\ww$,  let $\ww_{<t} = w_1 \cdots w_{t-1}$, and let $\ww_{>t} = w_{t+1} \cdots  w_{T}$. We construct $\widetilde{\ww}\defeq\ww_{<t}\mask\ww_{>t}$ by replacing $w_t$ in $\ww$ with a mask token $\mask$.
The alphabet is expanded to include $\mask$. 
We denote $\maskalphabet \defeq \alphabet \cup \{\mask\}$. \vspace{1mm}
The transformer $\HH^{(L)}$ is now of type $\maskalphabet^N\rightarrow \R^{D\times N}$.
A masked language gives the following probability distribution for position $t$
\begin{equation}
\begin{aligned}
    p(\widetilde{w}_t &\mid \ww_{<t}, \ww_{>t}) \\
    &= \softmax(\W[P]\HH^{(L)}(\widetilde{\ww})_{:,t} + \bb[P])_{\widetilde{w}_t}
\end{aligned}
\end{equation}
where $\widetilde{w}_t \in \maskalphabet$, $\W[P]\in\R^{|\maskalphabet| \times D}$ and $\bb[P]\in\R^{|\maskalphabet|}$.
In practice, multiple tokens may be masked and predicted simultaneously. 

\paragraph{Autoregressive Language Modeling.}
In contrast to masked language modeling, the goal of autoregressive language modeling (ALM) is to construct
a probability distribution over $\alphabet^*$.
To do so, the following factorization is employed
\begin{equation}
p(\ww) = p(\eos \mid \ww) \prod_{t=1}^T p(w_t \mid \ww_{<t})
\end{equation}
Every local conditional distribution $p(w_t \mid \ww_{<t})$ is defined over the set $\eosalphabet \defeq \alphabet \cup \{\eos\}$ and $\ww_{<1}\defeq \bos$, where $\bos,\eos\not\in\alphabet$.\footnote{This means that the transformer is a function of type $\underline{\alphabet}^{N+1} \rightarrow \R^{D \times (N+1)}$ where $\underline{\alphabet} \defeq \alphabet \cup \{\bos\}$.}
In a transformer-based autoregressive language model, each local conditional is parameterized as
\begin{equation}
\begin{aligned}
&    p(\overline{w}_t \mid \ww_{<t}) \\
    &\quad= \softmax(\W[P]\FF^{(L)}(\ww)_{:,t-1} + \bb[P])_{\overline{w}_t}
\end{aligned}
\end{equation}
where $\overline{w}_t \in \eosalphabet$, $\FF^{(L)}$ is a future-masked transformer, $\W[P]\in\R^{|\eosalphabet| \times D}$ and $\bb[P]\in\R^{|\eosalphabet|}$.

\section{A Transformer with Stack Attention}
\label{sec:method}
Recently \citet{deletang2023neural} showed that transformers are not able to learn several non-regular DCF languages.
Inspired by the fact that pushdown automata \citep{oettinger1961, SCHUTZENBERGER1963246}, automata that employ a single stack, can model CF languages \citep{10.1145/1463822.1463848}, we introduce a novel stack attention mechanism that emulates the functionality of a stack and integrate it into the transformer architecture, aiming to enable it to learn some CF languages.\looseness=-1

\subsection{Stacks over the Index Set}\label{sec:stack-def}
We first give a formal definition of a stack. 
In our paper, we define a stack as a data structure over the index set $[N]$. 
The state of a stack is a string $\stackseq\in[N]^*$ of indices.
There are three operations that we can perform that alter the state of the stack.
We describe each operation below in terms of $\stackseq$.
\begin{itemize}[leftmargin=*]
    \item The operation $\push \colon [N]^* \times [N] \rightarrow [N]^*$ adds an element to the top of the stack and is formally defined as follows:
    \begin{equation}
        \push(\stackseq, \stacksym) = \stackseq\stacksym
    \end{equation}
    \item The operation $\noop \colon [N]^* \rightarrow [N]^*$ leaves the stack unchanged and is defined as follows:
    \begin{equation}
        \noop(\stackseq) = \stackseq
    \end{equation}
    \item The operation $\pop \colon [N]^* \rightarrow [N]^*$ removes the top-most element of the stack and is formally defined as follows:
    \begin{subequations}
    \begin{align}
        &\pop(\varepsilon) = \varepsilon \\
        &\pop(\stacksym_1 \cdots \stacksym_{T}) = \stacksym_1 \cdots \stacksym_{T-1}     
    \end{align}
    \end{subequations}
\end{itemize}
We will use this definition in \Cref{thm:stack} to argue that our stack attention mechanism can be formally viewed as a type of stack.
Additionally, we will assume an operator $\peek : [N]^* \rightarrow \left([N] \cup \{0\}\right)$ 
that does not alter the state of the stack, but rather returns the top element (or $0$ if the stack is empty).
We define it below
\begin{subequations}
\begin{align}
  &\peek(\varepsilon) = 0 \\
  &\peek(\stacksym_1 \cdots \stacksym_{T}) = \stacksym_{T}     
\end{align}
\end{subequations}

\subsection{Stack Attention}\label{sec:stack-attention}
We now formally define our stack attention mechanism. 
We introduce a beginning-of-sequence symbol $\bos$ at the zeroth position, designated to represent an empty stack.
Each position $i \in \{0\}\cup[N]$ is assigned a distinct stack $\aalpha_{i}\in\R^{N+1}$. 
We write $\aalpha_i(j)$ to denote the $(j+1)^\text{th}$ value in $\aalpha_i$, for $0\leq j \leq N$. %
The stacks are defined inductively.
The initial stack, $\aalpha_0$, is constructed to attend to $\bos$ as follows
\begin{equation}
\label{eq:init}
    \aalpha_0 \defeq [1, 0, 0, \cdots]^{\top} \in \mathbb{R}^{N+1}
\end{equation}
Subsequent stacks are computed inductively based on the stack contents and the operations ($\push$, $\noop$, $\pop$) taken at previous timesteps.\footnote{We use the terms timestep and position interchangeably.} 
The three stack operations are defined for $i\geq1$ as follows:\looseness=-1
\begin{itemize}[leftmargin=*]
    \item $\push$ pushes the hidden state of the current position, so we just set the attention weight at the current position to be $1$ and the rest to be $0$, i.e.,
    \begin{align}
    \label{eq:push}
        \aalpha_{i}^{(\push)}(j) \defeq \begin{cases}
            1 & j = i \\
            0 & \text{otherwise}
            \end{cases} 
    \end{align}
    \item $\noop$ leaves the previous stack unchanged, so the stack from the last timestep is passed forward with no modification, i.e., we have
    \begin{equation}
        \aalpha_{i}^{(\noop)} \defeq \aalpha_{i-1}
    \end{equation}
    \item $\pop$ removes the top element, and backtracks to the second element in the stack, i.e., we have
    \begin{align}
    \label{eq:pop}
   \aalpha_{i}^{(\pop)} \defeq 
        \left[ \sum_{j=1}^{i-1} \aalpha_{i-1}(j) \aalpha_{j-1}\right] + \aalpha_{i-1}(0)\aalpha_0
    \end{align}
    The first term on the right-hand side retrieves the second element and is zeroed out when $i=1$. The second term accounts for the case of an empty stack---$\pop$ cannot be performed on an empty stack and in this case it is equivalent to a $\noop$.
\end{itemize}
The stack attention $\aalpha_{i}$ at position $i$ is computed as a superposition of the three operations:
\begin{equation}
\begin{aligned}
\label{eq:stack}
    \aalpha_{i} \defeq \aaa_i(\push)\cdot\,&\aalpha_{i}^{(\push)} + \aaa_i(\pop)\cdot \aalpha_{i}^{(\pop)} \\ 
     &+\aaa_i(\noop)\cdot \aalpha_{i}^{(\noop)}
\end{aligned}
\end{equation}
where $\aaa_i\in \Delta^2$ is a probability distribution over possible operations $\calA = \{\push,\pop,\noop\}$.
This distribution is determined by:
\begin{equation}
    \aaa_i \defeq \softmax \left(\W[A]\hh_i+\bb[A]\right)
\end{equation}
where $\W[A]\in\R^{3\times D}$ and $\bb[A]\in\R^3$ are learned parameters.

After obtaining the stack attention weights, we can compute the top element as a weighted sum just like standard self-attention:
\begin{equation}
    \Stack\left(\HH\right)_{:,i} \defeq \sum_{n=0}^{N} \aalpha_{i}(n) \hh_n
\end{equation}

\subsection{A Stack Transformer}
The stack is incorporated into the transformer by inserting a third sub-layer
in each transformer layer after the standard attention and feedforward layers defined in \cref{eq:translayer} and \cref{eq:translayerffn}:\looseness=-1
\begin{subequations}
\begin{align}
\HH^{(\ell)}_{\Stack} &\defeq \Stack\left(\HH^{(\ell)}_{\FFN}\right) + \HH^{(\ell)}_{\FFN} \\
    \HH^{(\ell)} &= \HH^{(\ell)}_{\Stack}
\end{align}
\end{subequations}
Similar to other sub-layers, we also employ a residual connection by summing the output of the stack attention mechanism $\Stack$ with its input, allowing the model to bypass the stack if needed. 
Layer normalization can also be used, but we omit due to initial results in preliminary experiments.
Because the rest of the model is left unchanged, it can be directly integrated into pre-trained language models to augment their ability to process hierarchical syntactic structures.

\begin{table*}[]
    \centering
    \begin{tabular}{lllllll}
        \toprule
        \multirow{2}{*}{Task} & \multicolumn{2}{c}{RNN} & \multicolumn{2}{c}{Transformer MLM} & \multicolumn{2}{c}{Transformer ALM} \\
         & Vanilla & Stack & Vanilla & Stack & Vanilla & Stack \\
        \midrule
        RS & $81.0 \pm 0.8$ & $100.0 \pm 0.0$ & $54.8 \pm 0.0$ & $100.0 \pm 0.0$ & $55.4 \pm 0.8$ & $100.0 \pm 0.0$ \\
        SM & $73.2 \pm 1.0$ & $100.0 \pm 0.0$ & $50.4 \pm 0.1$ & $93.1 \pm 4.4$ & $50.4 \pm 0.1$ & $92.8 \pm 2.6$ \\
        MA & $75.8 \pm 4.3$ & $91.0 \pm 6.3$ & $30.1 \pm 0.0$ & $34.3 \pm 1.4$ & $30.2 \pm 0.1$ & $29.5 \pm 0.6$ \\
        SE & $56.7 \pm 10.3$ & $89.9 \pm 7.2$ & $20.0 \pm 0.0$ & $29.8 \pm 8.0$ & $20.2 \pm 0.1$ & $20.3 \pm 0.2$ \\
        \bottomrule
    \end{tabular}
    \caption{Accuracies ($\%$) of the transformer and RNN without and with stacks on DCF tasks.}
    \label{tab:dcf}
    \end{table*}

\subsection{Computational Overhead}
\paragraph{Time.} The computation is bottlenecked by the $\pop$ operation, which sums over previous the previous positions and thereby has a time complexity of $\bigO{N}$. The total time complexity is $\bigO{N^2}$.
In contrast to standard attention, stack attention has to be computed sequentially, which breaks the parallelizability of the transformer and makes it substantially slower in practice. 
However, $\aaa$ and the output $\Stack(\HH)$ can still be computed in parallel.
Thus, $\aalpha$ is a function of the stack operations but \emph{not} of the hidden states.
It follows that if structural supervision of the stack operations is provided, e.g., as in \citet{sartran-etal-2022-transformer} and \citet{murty-etal-2023-pushdown}, $\aalpha_i$ for all $i \in [N]$
 can be pre-computed, and the entire model can be parallelized.\looseness=-1

\paragraph{Space.} The stack attention stores $N+1$ attentions of size $N$, so the space complexity is $\bigO{(N+1)N} = \bigO{N^2}$.
This is an improvement over the $\bigO{DN^2}$ space complexity of \citeposs{dusell2023stack} method.

\subsection{The Duality of Stack Attention}
Stack attention is both a stack over the index set, as defined in \Cref{sec:stack-def}, and an attention mechanism. 
In the following theorem, we make precise the manner in which 
our stack attention is a stack.

\paragraph{Notation.}
We use the symbol $\stackop_i$ to refer to an operation from the set $\{\push_i(\cdot), \noop(\cdot), \pop(\cdot)\}$
at every time step $i$. %
Note that $\push$, as defined in \Cref{sec:stack-def}, is a function of two arguments. 
However, we define $\push_i(\stackseq) \defeq \push(\stackseq, i)$, i.e., we push $i$, the index, to the stack. 
We introduce a function $\onehot$ of type $\{0\}\cup[N]\rightarrow \bool^{N+1}$ that converts an index into its one-hot encoding, a column vector with zeros everywhere except the given index, where the value is set to one. \looseness=-1
\begin{restatable}{theorem}{stack}
\label{thm:stack}
    Let $\stackop_1, \ldots, \stackop_N$ be a series of stack operations where $\stackop_i \in \{\push_i(\cdot), \noop(\cdot), \pop(\cdot)\}$ for all $i \in [N]$.
    Furthermore, suppose $\aaa_i(\stackop_i) = 1$ for all $i \in [N]$.
    Then, $\onehot[\peek(\stackop_{i}(\cdots \stackop_{1}(\varepsilon)))] = \aalpha_i$ for all $i \in \{0\} \cup [N]$.
\end{restatable}
\begin{proof}
    \cref{app:proof}
\end{proof}
Our stack-based attention is also an attention mechanism in the sense that it maintains a distribution over $\{0\}\cup[N]$.
We make this notion precise as well in the following theorem.
\begin{restatable}{theorem}{attention}
Consider a sequence of stack attention mechanisms $\aalpha_0, \ldots, \aalpha_N$.
Then, $\sum_{n=0}^{N}\aalpha_i(n)=1$ for all $i \in \{0\}\cup[N]$.
\end{restatable}
\begin{proof}
    \cref{app:proof}
\end{proof}

\subsection{Expressivity}
\label{sec:exp}
We leave the exact characterization of the expressivity of our stack-augmented transformer for future work.
However, we conjecture that it cannot model all the context-free languages \emph{without} positional encodings.
Such a result would mirror that of \citet{angluin2023masked}.\looseness=-1

To contextualize this conjecture, we first review the star-free languages. 
The star-free languages are regular languages definable by a regular expression \emph{without} Kleene star but with complement \citep{alma990008061930205503}.
They can also be characterized by finite-state automata with aperiodic transformation monoids \citep{SCHUTZENBERGER1965190}, also termed counter-free automata or permutation-free automata \citep{alma990008061930205503}. It has been shown that a counter-free automaton can only perform counting up to a threshold, but not modulo counting \citep{alma990008061930205503}.

Recently, \citet{angluin2023masked} showed that the class of languages recognizable by transformer encoders with hard attention, strict future masking, and no positional encodings, are exactly the star-free languages.
Building on this result, we conjecture that there exist non-star-free languages that are beyond the capability of a transformer encoder with (hard) stack attention and no positional encodings.
This conjecture is supported by our experiments in \cref{sec:res} where we show that stack-augmented transformers also fail to learn two tasks involving modulo counting. 
We hope to construct a proof of an expressivity result in future work.

\begin{table*}[t]
    \centering
    \begin{subtable}{\textwidth}
    \centering
    \begin{tabular}{lllllll}
        \toprule
        Task & Transformer & \none & \sincos & \relative & \rotary & \ALiBi   \\
        \midrule
        \multirow{2}{*}{RS} & Vanilla & $54.8 \pm 0.0$ & $50.7 \pm 0.2$ & $67.6 \pm 2.2$ & $55.4 \pm 1.2$ & $79.4 \pm 3.5$  \\
         & Stack & $100.0 \pm 0.0$ & $99.1 \pm 1.7$ & $100.0 \pm 0.0$ & $86.3 \pm 15.0$ & $100.0 \pm 0.0$ \\
         \midrule
        \multirow{2}{*}{SM} & Vanilla & $50.4 \pm 0.1$ & $49.5 \pm 0.6$ & $67.5 \pm 1.0$ & $52.1 \pm 1.8$ & $70.9 \pm 1.2$  \\
         & Stack & $93.1 \pm 4.4$ & $74.7 \pm 8.8$ & $98.5 \pm 1.1$ & $73.1 \pm 4.5$ & $92.9 \pm 2.7$ \\
         \midrule
         \multirow{2}{*}{MA} & Vanilla & $30.1 \pm 0.0$ & $30.1 \pm 0.0$ & $30.1 \pm 0.0$ & $30.1 \pm 0.0$ & $30.1 \pm 0.0$  \\
         & Stack & $34.3 \pm 1.4$ & $33.8 \pm 0.8$ & $35.0 \pm 1.1$ & $34.5 \pm 1.3$ & $34.7 \pm 1.1$ \\
         \midrule
         \multirow{2}{*}{SE} & Vanilla & $20.0 \pm 0.0$ & $20.0 \pm 0.0$ & $20.0 \pm 0.0$ & $20.0 \pm 0.0$ & $20.0 \pm 0.0$  \\
         & Stack & $29.8 \pm 8.0$ & $23.9 \pm 3.0$ & $25.2 \pm 1.8$ & $30.0 \pm 3.8$ & $27.9 \pm 5.8$ \\
        \bottomrule
    \end{tabular}
    \caption{MLM}
    \end{subtable}
    \begin{subtable}{\textwidth}
    \centering
    \begin{tabular}{lllllll}
        \toprule
        Task & Transformer & \none & \sincos & \relative & \rotary & \ALiBi   \\
        \midrule
        \multirow{2}{*}{RS} & Vanilla & $55.4 \pm 0.8$ & $55.2 \pm 0.7$ & $62.0 \pm 6.1$ & $72.9 \pm 3.5$ & $57.1 \pm 0.6$  \\
        & Stack & $100.0 \pm 0.0$ & $96.8 \pm 4.5$ & $100.0 \pm 0.0$ & $100.0 \pm 0.0$ & $100.0 \pm 0.0$ \\
        \midrule
        \multirow{2}{*}{SM} & Vanilla & $64.9 \pm 2.0$ & $60.8 \pm 3.1$ & $70.5 \pm 0.9$ & $71.9 \pm 0.9$ & $70.5 \pm 1.6$  \\
        & Stack & $92.8 \pm 2.6$ & $49.6 \pm 4.4$ & $93.2 \pm 2.3$ & $83.8 \pm 1.7$ & $93.4 \pm 1.2$ \\
        \midrule
        \multirow{2}{*}{MA} & Vanilla & $30.2 \pm 0.1$ & $25.7 \pm 2.3$ & $30.3 \pm 0.1$ & $26.0 \pm 0.8$ & $30.3 \pm 0.1$  \\
        & Stack & $30.0 \pm 0.1$ & $28.0 \pm 2.8$ & $30.3 \pm 0.3$ & $25.6 \pm 0.3$ & $30.3 \pm 0.1$ \\
        \midrule
        \multirow{2}{*}{SE} & Vanilla & $20.2 \pm 0.1$ & $20.2 \pm 0.3$ & $20.7 \pm 0.2$ & $20.3 \pm 0.2$ & $20.5 \pm 0.3$  \\
        & Stack & $20.3 \pm 0.2$ & $20.2 \pm 0.1$ & $20.7 \pm 0.1$ & $20.2 \pm 0.3$ & $20.3 \pm 0.1$ \\
        \bottomrule
    \end{tabular}
    \caption{ALM}
    \end{subtable}
    \caption{Performance comparison of a vanilla and stack transformer with different positional encodings.\looseness=-1}
    \label{tab:pe}
\end{table*}

\section{Deterministic CF Tasks}
We now discuss several tasks that are encodable by deterministic context-free grammars. 
\subsection{Tasks}
All four tasks we consider are derived from \citet{deletang2023neural} and are language transduction tasks. 
Every word from the input language $\xx\in \alphabet[I]^*$ is mapped to a word in the output language $\yy\in \alphabet[O]^*$ by means of a function $\ff \colon \alphabet[I]^* \rightarrow  \alphabet[O]^*$.
To convert a transduction task to a language acceptance task, a language is constructed over the alphabet $\alphabet=\alphabet[I]\cup\alphabet[O]$ as follows
\begin{equation}
   \Big\{ \xx \ff(\xx) \mid \xx \in \alphabet[I]^* \Big\} \subseteq \alphabet^*
\end{equation}
To experiment with this setup, in the case of an MLM, the input $\widetilde{\ww}$ is $\xx$ appended with $|\yy|$ mask tokens $\mask$. 
We then use the transformer to predict all the masked tokens at once and evaluate the predicted string $\yy'$ against $\yy=\ff(\xx)$.
Likewise, in the case of an ALM, given a prefix $\xx$, we sample $y'_t \sim p(\cdot \mid \xx\yy'_{<t})$ autoregressively, where $y'_t$ denotes the $t^{\text{th}}$ symbol of $\yy'$ and $\yy'_{<t}=y'_1\cdots y'_{t-1}$.
As in the case of MLM, we evaluate the predicted $\yy'$ against the $\yy = \ff(\xx)$. We follow the choices of \citet{deletang2023neural} for $\alphabet[I]$ and $\alphabet[O]$

\paragraph{Reverse String (RS).} 
The first task is to compute the reverse of an input string, i.e., $\ff[RS](\xx)=\xx^R$. 
In this task, we take $\alphabet[I]=\alphabet[O]=\{\syma, \symb\}$.%
We give an example below.
\begin{testexample}
    \begin{align*}
        \xx &= \syma \symb \symb \\
        \yy &= \symb \symb \syma 
    \end{align*}
\end{testexample}

\paragraph{Stack Manipulation (SM).} 
In the second task, the input string $\xx$ consists of a stack of two symbols $\{\syma,\symb\}$,
printed from bottom to top, and a sequence of stack operations drawn from the set $\{\tpusha, \tpushb, \tpop\}$. 
The function $\ff[SM](\xx)$ outputs the final stack after all the given operations are executed sequentially on the input stack, printed from top to bottom. 
We always have $|\yy|=|\xx|+1$. 
If the final stack has fewer elements than $|\yy|$, it will be padded with $\pad$ tokens, which are ignored when accuracy is computed. 
We have $\alphabet[I]=\{\syma,\symb,\tpusha,\tpushb,\tpop\}$ and $\alphabet[O]=\{\syma,\symb,\pad\}$.
An example is given below. 
\begin{testexample}
\begin{align*}
    \xx &= \symb \syma \symb \tpop \tpusha \tpushb \\
    \yy &= \symb \syma \syma \symb \pad \pad \pad
\end{align*}
\end{testexample}

\paragraph{Modular Arithmetic (MA).} 
In the third task, we consider a transduction task based on modular arithmetic. 
An algebraic expression $\xx$ consists of five numerical constants $\{0,1,2,3,4\}$, three operations $\{+,-,\cdot\}$, brackets $\{(,)\}$, and a congruence sign $\{\equiv\}$. We say two integers are congruent if and only if a pre-set modulus is a divisor of their difference. In this task, we set the modulus to $5$, so the function $\ff[MA]$ evaluates the expression modulo $5$.
We have $\alphabet[I]=\{0,1,2,3,4,+,-,\cdot,(,),\equiv\}$ and $\alphabet[O]=\{0,1,2,3,4\}$.
An example is given below.
\begin{testexample}
    \begin{align*}
        \xx &= (1 + 2) \cdot 3 \equiv\\
        \yy &= 4
    \end{align*}
\end{testexample}%

\paragraph{Solve Equation (SE).}
In our fourth and final task, we consider a transduction task that solves equations over a single variable, which we denote $z$.
The input string $\xx$ is a modular equation with five constants $\{0,1,2,3,4\}$, two operations $\{+,-\}$, brackets $\{(,)\}$, a congruence sign $\{\equiv\}$, and a variable $\{z\}$. The modulus is set to $5$. 
The function $\ff[SE]$ solves this equation and returns the value of the variable. 
We have $\alphabet[I]=\{0,1,2,3,4,+,-,(,),\equiv,z\}$ and $\alphabet[O]=\{0,1,2,3,4\}$. 
An example is given below.\looseness=-1
\begin{testexample}
    \begin{align*}
        \xx &= (1 + z) + 2 \equiv 2 \\
        \yy &= 4
    \end{align*}
\end{testexample}
\subsection{Experimental Setup}
\label{sec:setup}
Following \citet{deletang2023neural}, we experiment with a transformer with the number of layers $L=5$ and the model size $D=64$.
Unless otherwise specified, no positional encodings are used. We discuss the effect of various positional encodings in \cref{sec:pe}. 
Length generalization has been the focus of many papers in this line of research \citep{NIPS2015_26657d5f, deletang2023neural}. We follow suit to train on input strings $\xx$ with $1\leq |\xx|\leq 40$ and test on $\xx$ with $40< |\xx|\leq 100$.
Training details can be found in \cref{app:setup}.\looseness=-1

\subsection{Results}
\label{sec:res}
We report our results of the four DCF tasks in \cref{tab:dcf} and \cref{tab:pe}.

\subsubsection{Transformer vs. Stack Transformer}
We report the performance of the standard transformer and our stack-augmented transformer on the four DCF tasks presented in \cref{tab:dcf}. 
For comparison, we also exhibit results of vanilla recurrent neural networks (RNNs) and stack-RNNs \citep{NIPS2015_26657d5f}. 
As expected, the vanilla transformer exhibits poor performance on all the DCF tasks. 
After being augmented with a stack, the transformer improves from nearly chance to over $90\%$ on RS and SM. 
These results demonstrate that our stack-augmented attention helps on some tasks.
However, on MA and SE, the performance after adding the stack attention only improves slightly; it still falls far behind stack RNNs and even vanilla RNNs.
We conjecture that the reason for this shortcoming is our stack transformer's incapability to learn non-counter-free languages---both the last two tasks (MA) and SE require the ability to perform modular arithmetic, which makes them non-star-free, as discussed in \cref{sec:exp}. 
Additionally, \citet{feng2023towards} also directly prove that the transformer cannot perform modular arithmetic.\looseness=-1

\subsubsection{Positional Encodings}
\label{sec:pe}
In this section, we add various positional encodings to the transformer and investigate their effect.
We consider five different positional encodings: $\none$, $\sincos$, $\relative$, $\rotary$, and $\ALiBi$; see \cref{app:pe} for more details.
As our stack attention is computed inductively, 
positional information is already present in the model, reducing the need for positional encodings. 
This is evident in \cref{tab:pe}, where including positional encodings generally has a \emph{negative} impact on the stack transformer's performance.
Most notably, $\sincos$ and $\rotary$ heavily degrade stack transformer's
performance on RS and SM. 
However, $\relative$ constitutes an exception, as it results in improved performance on SM. 
In contrast, with the standard transformer architecture, positional encodings \emph{do} seem to help on star-free tasks. 
The largest improvement comes from $\ALiBi$ in the MLM setting and $\rotary$ in the ALM setting. 
Nevertheless, none of the investigated positional encodings are able to boost the performance of vanilla transformer to anywhere near that of our stack-augmented transformer.\looseness=-1

\begin{table*}[]
    \centering
    \begin{tabular}{llllll}
        \toprule
        \multirow{2}{*}{Model} & \multirow{2}{*}{Task} & \multicolumn{2}{c}{Penn Treebank} & \multicolumn{2}{c}{WikiText-2} \\
         & & Vanilla & Stack & Vanilla & Stack \\
        \midrule
        \multirow{2}{*}{Scratch} & MLM & $95.53 \pm 19.66$ & $34.28 \pm 2.76$ & $73.74 \pm 3.79$ & $64.75 \pm 1.75$ \\
        \cmidrule{2-6}
         & ALM & $73.14 \pm 0.34$ & $69.86 \pm 0.26$ & $191.01 \pm 0.71$ & $206.42 \pm 0.80$ \\
        \midrule
        \multirow{2}{*}{Pretrained} & MLM & $3.99\pm 0.08 $ & $4.46\pm 0.11$ & $4.41 \pm 0.12$ & $4.65 \pm 0.06$ \\
        \cmidrule{2-6}
         & ALM & $21.26 \pm 0.03$ & $32.36 \pm 0.16$ & $29.29 \pm 0.02$ & $54.96 \pm 0.19$ \\
        \bottomrule
    \end{tabular}
    \caption{MLM and ALM Perplexities on WikiText-2 and PTB.}
    \label{tab:lm}
    \vspace{-10pt}
\end{table*}

\section{Language Modeling}
\label{sec:res_lm}
We consider masked language modeling using RoBERTa \cite{liu2020roberta} and autoregressive language modeling using  GPT-2 \citep{radford2019language}. 
Following the experimental setup proposed by previous authors \citep{NIPS2015_26657d5f, dusell2023stack}, we experiment on the Penn Treebank (PTB), licensed through the LDC \citep{marcus-etal-1994-penn}, and WikiText-2 \cite{merity2017pointer}. 
We consider models both trained from scratch and fine-tuned from pre-trained weights. 
The pre-trained models and datasets are obtained from HuggingFace \citep{wolf-etal-2020-transformers, lhoest-etal-2021-datasets}. 
See \cref{app:setup} for more details about setup and hyperparameters.\looseness=-1

\label{sec:res_lm}
The results in \cref{tab:lm} are mixed. 
Our major finding is that transformers benefit from the stack attention when training data is scarce, but the benefits gradually diminish as the size of training data grows.
More concretely, when the models are trained from scratch, the addition of our stack attention mechanism \emph{does} result in a noticeable benefit in most settings.
In the MLM setting, where $15\%$ of the tokens are replaced with $\mask$, stacks reduce the perplexity under the trained model on the held-out split from $95.53$ to $34.28$ on PTB and from $73.74$ to $65.22$ on WikiText-2.
In the ALM setting, the stack transformer still slightly improves the performance on PTB---perplexity drops from $73.14$ to $69.86$.
However, the stack transformer is less effective on WikiText-2, whose training set is larger. 
Moreover, when we fine-tune from pre-trained models, stacks are always detrimental across the two datasets in both MLM and ALM settings.\looseness=-1

\section{Discussion}
From the results described in \cref{sec:res} and \cref{sec:res_lm}, we observe two trends:
\begin{itemize}[leftmargin=*]
    \item The positive impact of stack attention is evident on \citeposs{deletang2023neural} 4 DCF tasks (especially on 2 of the 4), but almost nonexistent on English language modeling;
    \item On the English language modeling task, stack attention is more helpful in settings with limited training data, but is less helpful and can even be harmful when the model is trained on a larger amount of data.\looseness=-1
\end{itemize}
We interpret these trends as support for the idea that stack attention improves the representational capacity of a transformer language model and, additionally, confers an inductive bias to the transformer architecture that allows it to better learn certain context-free tasks more efficiently.
The larger representational capacity explains why the performance on certain tasks, i.e., RS and SM, improves drastically with the addition of stack attention and the better inductive bias explains why transformer language models with stack attention perform better with less training data on the English modeling task.
However, the fact that a vanilla transformer language model performs on par with stack attention when modeling larger English language datasets suggests that a good inductive bias is not needed for larger data sets.
This suggests that, in contrast to the viewpoint of traditional linguistic theory \cite{Chomsky+1957}, models that are higher on the Chomsky hierarchy are \emph{not} necessary for developing a good statistical language model.
We believe this claim is consistent with the literature, in which few successful large language models are endowed with a syntactic bias. 
However, there are many smaller syntax-infused language models \citep{dyer-etal-2016-recurrent} that do work well on smaller data, as ours does.\looseness=-1

\section{Conclusion}
We propose a novel implementation of a differentiable stack and show that a transformer augmented with such stacks can solve certain deterministic context-free tasks that are beyond the capability of standard transformers. However, unlike a stack RNN, the stack transformer cannot model the entire class of deterministic context-free languages. 

\section*{Acknowledgements}
This publication was made possible by an ETH AI Center doctoral fellowship to Jiaoda Li.
Ryan Cotterell acknowledges support from the Swiss National Science
Foundation (SNSF) as part of the ``The Forgotten Role of Inductive Bias in Interpretability'' project.

\section*{Limitations}
The primary limitation of the proposed stack attention is it only allows one $\pop$ operation at a time. 
It can be extended to have multiple $\pop$s in a manner similar to \citet{yogatama2018memory}. It can also only handle deterministic context-free languages. We would like to extend it to non-deterministic stacks in future works. 
Although our method does not require structural supervision, it can in principle take advantage of it when it is available. In such cases, the model can be fully parallelized, leading to great improvement in time efficiency. It would be interesting to explore this possibility in the future.\looseness=-1

\section*{Ethical Considerations}
We foresee no ethical concerns with this work.

\bibliography{custom}

\appendix
\onecolumn
\section{Proof}
\label{app:proof}
\stack*
\begin{proof}
    \item
    \paragraph{Base case ($i=0$).} The stack is initialized to be empty, i.e., $\stackseq_0=\varepsilon$ and $\peek(\stackseq_0)=0$.
    By definition, we have
    \begin{equation}
        \aalpha_0 = [1, 0, \ldots]^\top = \onehot[\peek(\stackseq_0)]
    \end{equation}

    \paragraph{Inductive Step.} 
    Suppose there exists an $i>0$, such that $\forall i'< i$, $\aalpha_{i'}=\onehot[\peek(\stackseq_{i'})]$, and $\aaa_i(\stackop_i)=1$.
    \begin{itemize}[leftmargin=*]
        \item If $\stackop_i=\push$, $\stackseq_i=\push(\stackseq_{i-1})=\stackseq_{i-1} i$ and $\peek(\stackseq_i)=i$, so according to \cref{eq:push} we have $\aalpha_i=\onehot[\peek(\stackseq_i)]$. 
        \item If $\stackop_i=\noop$, $\stackseq_i=\noop(\stackseq_{i-1})=\stackseq_{i-1}$, and $\aalpha_{i}=\aalpha_{i-1}$. Since $\aalpha_{i-1}=\onehot[\peek(\stackseq_{i-1})]$, we have $\aalpha_{i}=\onehot[\peek(\stackseq_i)]$. 
        \item If $\stackop_i=\pop$,  
        \begin{align}
            \aalpha_i &= \sum_{j=1}^{i-1} \aalpha_{i-1}(j) \aalpha_{j-1} + \aalpha_{i-1}(0)\aalpha_0
        \end{align}
        If $\aalpha_{i-1}(0)=1$, i.e. $\stackseq_{i-1}$ is empty, $\stackseq_i=\pop(\varepsilon)=\varepsilon$, and 
        \begin{subequations}
            \begin{align}
                \aalpha_i &= \aalpha_{i-1}(0)\aalpha_0 \\
                &= \aalpha_0 \\
                &= \onehot[0] \\
                &= \onehot[\peek(\stackseq_i)]
            \end{align}     
        \end{subequations}

        Otherwise, 
        \begin{subequations}
            \begin{align}
                \aalpha_i &= \sum_{j=1}^{i-1} \aalpha_{i-1}(j) \aalpha_{j-1}  \\
                &=\aalpha_{\peek(\stackseq_{i-1})-1} \\
                &=\onehot[\peek(\stackseq_{\peek(\stackseq_{i-1})-1})] \\
                &=\onehot[\peek(\pop(\stackseq_{i-1}))] \label{eq:pop_proof}\\
                &=\onehot[\peek(\stackseq_i)]
            \end{align}           
        \end{subequations}

        One can understand \cref{eq:pop_proof} intuitively as follows: $\stackseq_{\peek(\stackseq_{i-1})-1}$ is the stack right before the current stack top $\peek(\stackseq_{i-1})$ is pushed, so the stack top at $\peek(\stackseq_{i-1})-1$ is the second top-most element at $i-1$, i.e., $\peek(\stackseq_{\peek(\stackseq_{i-1})-1})=\pop(\stackseq_{i-1})$.
    \end{itemize}
\end{proof}

\attention*
\begin{proof}
    \item
    \paragraph{Base case} It holds for $\aalpha_0 = [1, 0, \ldots]^\top$.
    \paragraph{Induction step} Suppose there exists an $i> 0$, such that $\forall i'< i$, $\sum_{n=0}^N\aalpha_{i'}(n)=1$.
    \begin{itemize}[leftmargin=*]
        \item $\push$. Obviously, 
        \begin{equation}
            \sum_{n=0}^{N}\aalpha_{i}^{(\push)}(n)=1
        \end{equation}
        \item $\noop$. Since $\aalpha_{i}^{(\noop)} = \aalpha_{i-1}$, we also have
        \begin{equation}
            \sum_{n=0}^{N}\aalpha_{i}^{(\noop)}(n)=1
        \end{equation}
        \item $\pop$.     
        \begin{subequations}
            \begin{align}
                \sum_{n=0}^{N}\aalpha_{i}^{(\pop)}(j) &= \sum_{n=0}^{N}\left(\sum_{j=1}^{i-1} \aalpha_{i-1}(j) \aalpha_{j-1} + \aalpha_{i-1}(0)\aalpha_0\right)(n) \\
                &= \sum_{n=0}^{N}\left(\sum_{j=1}^{i-1} \aalpha_{i-1}(j) \aalpha_{j-1}(n) + \aalpha_{i-1}(0)\aalpha_0(n)\right) \\
                &= \sum_{j=1}^{i-1} \aalpha_{i-1}(j) \left(\sum_{n=0}^{N}\aalpha_{j-1}(n)\right) + \aalpha_{i-1}(0)\left(\sum_{n=0}^{N}\aalpha_0(n)\right) \\
                &= \sum_{j=1}^{i-1} \aalpha_{i-1}(j) +  \aalpha_{i-1}(0) \\
                &= \sum_{j=0}^{i-1} \aalpha_{i-1}(j) \\
                &= 1
             \end{align}
        \end{subequations}

    \end{itemize}

    Therefore, 
    \begin{subequations}
        \begin{align}
            &\sum_{n=0}^{N} \aalpha_{i}(n) \\
            = &\sum_{n=0}^{N} \left(\sum_{a\in\calA} \aaa_i(a) \aalpha_{i}^{(a)}\right)(n) \\
            =&\sum_{a\in\calA}\aaa_i(a)\sum_{n=0}^{N}\aalpha_{i}^{(a)}(n) \\
            =&\sum_{a\in\calA}\aaa_i(a) \\
            =& 1
        \end{align}
    \end{subequations}

\end{proof}

\section{Experimental Setup}
\label{app:setup}
\subsection{DCF Tasks}
The model is trained using the Adam optimizer \cite{DBLP:journals/corr/KingmaB14} with a learning rate of $1e^{-4}$, which we find works well for all the tasks.
On the RS and SM tasks, we use a batch size of 32 and we train the model for $100,000$ steps. 
On the MA and SE tasks, the batch size and the number of training steps are increased to 128 and $1,000,000$, respectively, to ensure sufficient training. 
Each experiment is run 5 times with different random seeds. Means and variances of accuracies are reported \cref{tab:dcf} and \cref{tab:pe}. %

\subsection{Language Modeling}
The texts in the datasets
are grouped into chunks of 128 tokens. 
Each model is, again, trained using the Adam optimizer for a maximum of 100 epochs with early stopping applied when the validation loss has not improved for 5 epochs in a row. We tune the learning rate from $\{1e^{-5}, 2e^{-5}, 1e^{-4}, 2e^{-4}\}$ on the validation set, and choose $2e^{-5}$ that leads to the best validation performance.
Results on the test set over 5 runs with different random seeds are reported in \cref{tab:lm}. 
Experiments are conducted on a single NVIDIA Tesla V100 GPU.

\section{Positional Encodings}
\label{app:pe}
We consider five different commonly used positional encodings:
\begin{itemize}
    \item $\none$. No positional encodings are used.
    \item $\sincos$. The sinusoidal positional encodings used in vanilla transformer \cite{NIPS2017_3f5ee243}. Positional information encoded sinusoidally is added to the embeddings. 
    \item $\relative$. In Transformer-XL \cite{dai-etal-2019-transformer}, relative rather than absolute sinusoidal positional information is added to the keys and queries of each attention block.
    \item $\rotary$. Introduced by \citet{DBLP:journals/corr/abs-2104-09864} and popularized by GPT-3 \cite{NEURIPS2020_1457c0d6}, rotary positional encodings multiply the keys and queries by sinusoidal encodings.
    \item $\ALiBi$. \citet{press2022train} adds linear biases to the attention blocks that favor the more recent tokens. 
\end{itemize} 

\section{Analysis: Attention Maps}
\label{app:map}
An advantage of our stack attention mechanism is that we can visualize the stack tops $\aalpha_i$, which provides greater interpretability than methods where stack tops are mixtures of hidden states \cite{NIPS2015_26657d5f}. %
We run a set of toy experiments with the stack transformer in the MLM setting. We randomly select one test example for each task. 

\paragraph{RS.} 
At the first two layers (\cref{fig:reverse1}, \cref{fig:reverse2}), the first $5$ tokens attend to themselves while the $\mask$ tokens attend to the last token in $\xx$. The most probable sequence of operations that leads to such a stack attention map is the input $\xx$ is pushed one by one onto the stack and $\noop$ is performed on all the $\mask$ tokens.
At the third layer (\cref{fig:reverse3}), the stacks for the $\mask$ tokens shift one position backward at a time, which demonstrates the stack elements are $\pop$ed one by one to generate the outputs. %
At the last two layers, all the tokens attend to themselves, so the stacks can be regarded as being skipped (\cref{fig:reverse4}, \cref{fig:reverse5}).%

\paragraph{SM.} Looking at the attention map at the first layer (\cref{fig:stack1}), we can infer the operations taken by the stack as follows: the stack first $\push$es the initial stack contents ($\syma\symb$);
once the $\tpop$ operation is read, it reverts to the first element $\syma$; 
then it performs the operation $\tpushb$ twice as instructed;
afterwards, it $\pop$s the final stack $\symb\symb\syma$ for outputs. 
The stack attention correctly skips the $\symb$ at timestep 1 as it has already been $\pop$ed at timestep 2. 
The last three positions are $\pad$ tokens and can be ignored. 

\paragraph{MA and SE.}
We also provide an attention map for MA and SE in \cref{fig:MA} and \cref{fig:SE}. 
Their attention maps are less interpretable as the stack transformer does not learn them well. Nevertheless, we can still observe that the stacks seem to be able to match the parentheses, which matches our expectations of the stack's strengths. For MA, at the first layer (\cref{fig:modular1}), the stack successfully matches the last two closing parentheses (at timestep 8 and 9) with their corresponding open parentheses (at timestep 5 and 0 respectively). For SE, the pattern is less obvious presumably because the parentheses do not have an impact on the order of arithmetic operations and can be ignored.

\begin{figure*}[htbp]
    \centering
    \begin{subfigure}{.32\textwidth}
        \centering
        \includegraphics{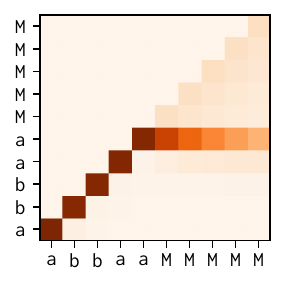}
        \caption{Layer 1}
        \label{fig:reverse1}
    \end{subfigure}
    \begin{subfigure}{.32\textwidth}
        \centering
        \includegraphics{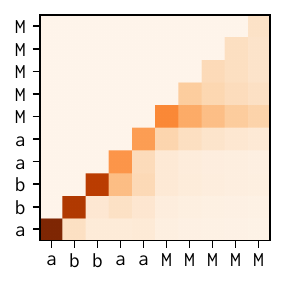}
        \caption{Layer 2}
        \label{fig:reverse2}
    \end{subfigure}
    \begin{subfigure}{.32\textwidth}
        \centering
        \includegraphics{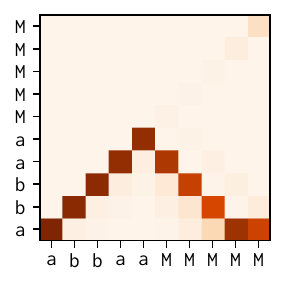}
        \caption{Layer 3}
        \label{fig:reverse3}
    \end{subfigure}
    \begin{subfigure}{.32\textwidth}
        \centering
        \includegraphics{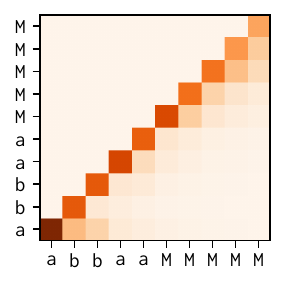}
        \caption{Layer 4}
        \label{fig:reverse4}
    \end{subfigure}
    \begin{subfigure}{.32\textwidth}
        \centering
        \includegraphics{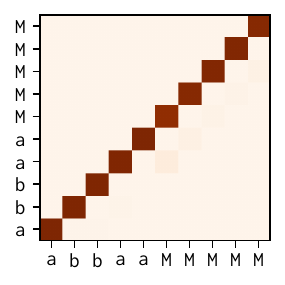}
        \caption{Layer 5}
        \label{fig:reverse5}
    \end{subfigure}
    \caption{Stack attention maps at different layers for RS. The input $\xx$ is $\syma \symb \symb \syma \syma$. $\symM$ represents a $\mask$ token.}
    \label{fig:RS}
\end{figure*}

\begin{figure*}[htbp]
    \centering
    \begin{subfigure}{.32\textwidth}
        \centering
        \includegraphics{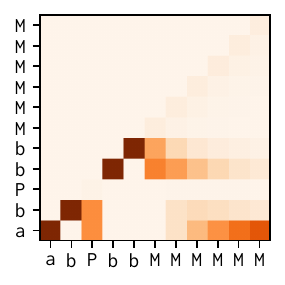}
        \caption{Layer 1}
        \label{fig:stack1}
    \end{subfigure}
    \begin{subfigure}{.32\textwidth}
        \centering
        \includegraphics{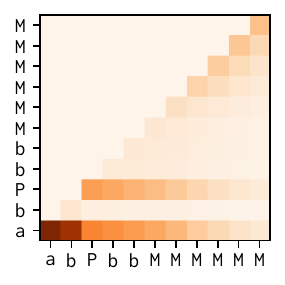}
        \caption{Layer 2}
        \label{fig:stack2}
    \end{subfigure}
    \begin{subfigure}{.32\textwidth}
        \centering
        \includegraphics{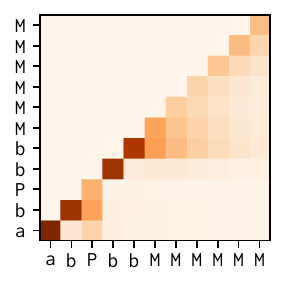}
        \caption{Layer 3}
        \label{fig:stack3}
    \end{subfigure}
    \begin{subfigure}{.32\textwidth}
        \centering
        \includegraphics{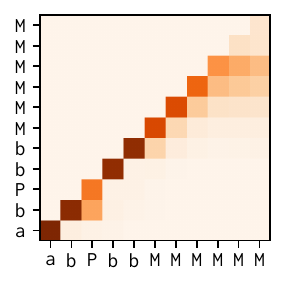}
        \caption{Layer 4}
        \label{fig:stack4}
    \end{subfigure}
    \begin{subfigure}{.32\textwidth}
        \centering
        \includegraphics{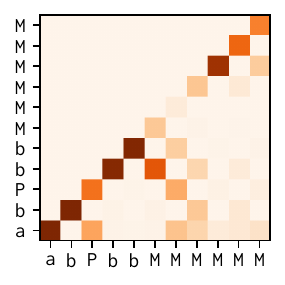}
        \caption{Layer 5}
        \label{fig:stack5}
    \end{subfigure}
    \caption{Stack attention maps at different layers for SM. The input $\xx$ is $\syma \symb \tpop \tpusha \tpushb$. In the graphs, $\tpusha$, $\tpushb$, and $\tpop$ are abbreviated as $\syma$, $\symb$, and $\symP$ respectively. 
    $\symM$ represents a $\mask$ token. The correct output should be $\symb \symb \syma$ followed by $\pad$ tokens.}
    \label{fig:SM}
\end{figure*}

\begin{figure*}[htbp]
    \centering
    \begin{subfigure}{.32\textwidth}
        \centering
        \includegraphics{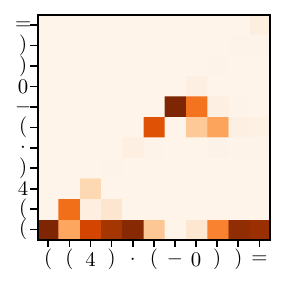}
        \caption{Layer 1}
        \label{fig:modular1}
    \end{subfigure}
    \begin{subfigure}{.32\textwidth}
        \centering
        \includegraphics{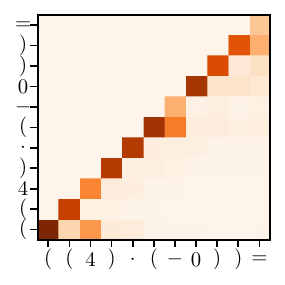}
        \caption{Layer 2}
        \label{fig:modular2}
    \end{subfigure}
    \begin{subfigure}{.32\textwidth}
        \centering
        \includegraphics{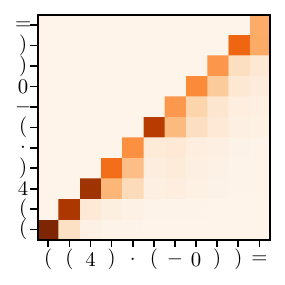}
        \caption{Layer 3}
        \label{fig:modular3}
    \end{subfigure}
    \begin{subfigure}{.32\textwidth}
        \centering
        \includegraphics{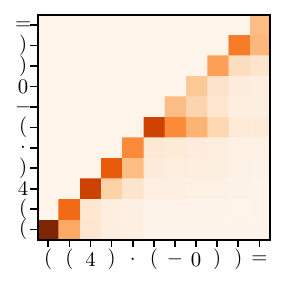}
        \caption{Layer 4}
        \label{fig:modular4}
    \end{subfigure}
    \begin{subfigure}{.32\textwidth}
        \centering
        \includegraphics{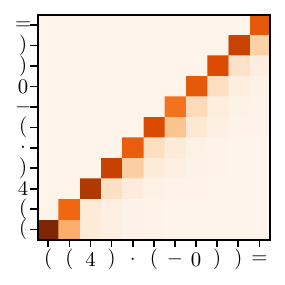}
        \caption{Layer 5}
        \label{fig:modular5}
    \end{subfigure}
    \caption{Stack attention maps at different layers for MA. The input $\xx$ is $((4)\cdot(-0))=$.}
    \label{fig:MA}
\end{figure*}

\begin{figure*}[htbp]
    \centering
    \begin{subfigure}{.32\textwidth}
        \centering
        \includegraphics{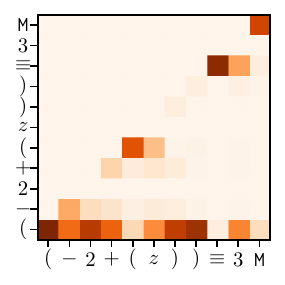}
        \caption{Layer 1}
        \label{fig:solve1}
    \end{subfigure}
    \begin{subfigure}{.32\textwidth}
        \centering
        \includegraphics{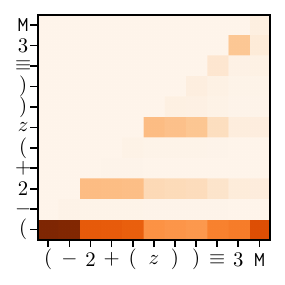}
        \caption{Layer 2}
        \label{fig:solve2}
    \end{subfigure}
    \begin{subfigure}{.32\textwidth}
        \centering
        \includegraphics{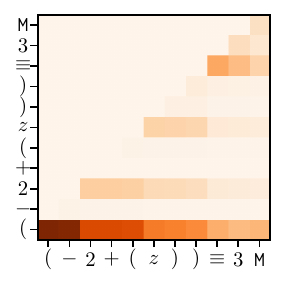}
        \caption{Layer 3}
        \label{fig:solve3}
    \end{subfigure}
    \begin{subfigure}{.32\textwidth}
        \centering
        \includegraphics{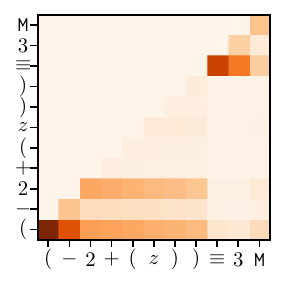}
        \caption{Layer 4}
        \label{fig:solve4}
    \end{subfigure}
    \begin{subfigure}{.32\textwidth}
        \centering
        \includegraphics{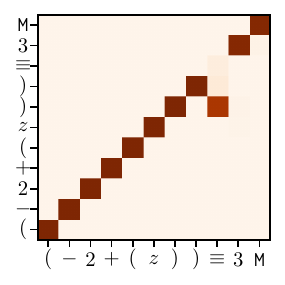}
        \caption{Layer 5}
        \label{fig:solve5}
    \end{subfigure}
    \caption{Stack attention maps at different layers for SE. The input $\xx$ is $(1+(-z))=3$. $\symM$ represents a $\mask$ token.}
    \label{fig:SE}
\end{figure*}
\newpage
\section{Related Work}
\subsection{Stack Augmentation}
Equipping a neural network with a data structure such as an external stack to enhance its ability to recognize context-free languages has been extensively investigated in previous works \cite{10.1007/BF00114845, das1992learning, NIPS1992_45645a27, 279194}.
The idea has seen a resurgence in recent years, with work focusing primarily on recurrent networks \cite{NIPS2015_26657d5f, NIPS2015_b9d487a3, hao-etal-2018-context, yogatama2018memory, DBLP:journals/corr/abs-1911-03329, dusell-chiang-2020-learning, dusell2022learning}. 
\citet{NIPS2015_26657d5f} propose to superpose the result of applying each stack operation at each step, which directly inspires our work. 
We adapt it for application to transformers by rendering this concept as an attention mechanism. 
In that sense, our work is related to \citet{das1992learning} and \citet{NIPS2015_b9d487a3}, which also assign weights to stack elements. 
Our stack attention mechanism is different as the stack attention weights are assigned to previously seen tokens indicating where the top element is located

\citet{sartran-etal-2022-transformer} and \citet{murty-etal-2023-pushdown} incorporate a stack mechanism into a transformer language model with structural supervision during training.
\citeposs{dusell2023stack} contemporaneous work also augments a transformer language model with a stack. Both their and our methods are named stack attention, but their stack attention is an attention mechanism over stack actions while ours is an attention mechanism over input tokens. 

\subsection{Expressivity of Transformers}
The expressivity of transformers under various assumptions has been extensively studied. A stream of research considers transformer encoders with a classification layer at the end as recognizers. \citet{hahn-2020-theoretical} proves that transformers cannot recognize parity language, a periodic language of binary strings with an even number of $1$'s, and Dyck-2 language, a CF language of balanced brackets 
of two types. \citet{bhattamishra-etal-2020-ability} find that transformers can recognize certain counter languages but fail to recognize non-star-free languages such as $(\syma\syma)^*$. \citet{svete-cotterell-2024-transofrmers} show that transformers can represent $n$-gram language models. \citet{hao-etal-2022-formal}, \citet{10.5555/3618408.3618629}, \citet{NEURIPS2023_a48e5877}, \citet{barcelo2024logical}, and  \citet{angluin2023masked} relate transformers to circuit complexity and formal logic. With various extensions, transformers' expressivity can be increased. \citet{pmlr-v139-weiss21a} propose a programming language that shares the same basic operations with transformers but is more expressive than standard transformers. \citet{JMLR:v22:20-302} and \citet{merrill2024the} show that transformer encoder--decoders and decoders are Turing complete with additional scratch space. 

\end{document}